\documentclass[letterpaper, 10 pt, conference]{style/ieeeconf}

\IEEEoverridecommandlockouts                              

\usepackage[pdftex]{graphicx}
\usepackage{booktabs}
\usepackage{moreverb}
\usepackage{url}
\usepackage{amsmath,amssymb}
\usepackage{multicol,lipsum}
\usepackage{bm}
\usepackage{color, colortbl}
\usepackage[colorlinks,bookmarksopen,bookmarksnumbered,citecolor=red,urlcolor=red]{hyperref}
\usepackage{svg}
\usepackage{multirow}

\hypersetup
{
	pdftitle = {Whole-body control for quadrupedal locomotion on challenging terrain},
	pdfauthor = {Shamel Fahmi},
	pdfsubject = {RA-L manuscript},
	pdfkeywords = {whole-body control, legged robots, challenging locomotion, and
	terrain mapping},
	pdftoolbar = true,
	colorlinks = true,
	linkcolor = black,
	citecolor = black,
	urlcolor = black,
}

\usepackage{glossaries}
\usepackage[tight]{units}
\usepackage[normalem]{ulem} 
\usepackage{cancel}

\newacronym{com}{CoM}{Center of Mass}
\newacronym{cop}{CoP}{Center of Pressure}
\newacronym{zmp}{ZMP}{Zero Moment Point}
\newacronym{fddp}{FDDP}{Feasible Differential Dynamic Programming}
\newacronym{cp}{CP}{Capture Point}
\newacronym{grfs}{GRFs}{Ground Reaction Forces}
\newacronym{vhsip}{VHSIP}{Variable Height Springy Inverted Pendulum}
\newacronym{ik}{IK}{Inverse Kinematic}
\newacronym{ocp}{OCP}{Optimal Control Problem}
\newacronym{nlp}{NLP}{Nonlinear Programming}
\newacronym{ltv}{LTV}{Linear Time Varying}
\newacronym{lti}{LTI}{Linear Time Invariant}
\newacronym{td}{TD}{Touch Down}
\newacronym{lc}{LC}{Landing Controller}
\newacronym{qp}{QP}{Quadratic Program}
\newacronym{rl}{RL}{Reinforcement Learning}
\newacronym{mpc}{MPC}{Model Predictive Control}
\newacronym{fwp}{FWP}{Feasible Wrench Polytope}
\newacronym{micp}{MICP}{Mixed-Integer Convex Program}
\newacronym{srbd}{SRBD}{Single Rigid Body Dynamics}
\newacronym{to}{TO}{Trajectory Optimization}
\newacronym{ddp}{DDP}{Differential Dynamic Program}

\newacronym{dofs}{DoFs}{Degrees of Freedom}
\newacronym{ode}{ODE}{Ordinary Differential Equation}
\newacronym{msd}{MSD}{Mass-Spring-Damper}
\newacronym{pwbc}{pWBC}{projection-based Whole Body Control}
\newacronym{sp}{SP}{Support Polygon}
\newacronym{imu}{IMU}{Inertial Measurement Unit}

\newacronym{rt}{RT}{Real Time}

\newacronym{slip}{SLIP}{Spring Loaded Inverted Pendulum}
\newacronym{eom}{EoM}{Equation of Motions}
 
\newacronym{sqp}{SQP}{Sequential Quadratic Programming}
\newacronym{mic}{MIC}{Mixed-Integer Convex}
\newacronym{cmaes}{CMA-ES}{Covariance Matrix Adaptation Evolution Strategy}
\newacronym{ara}{ARA*}{Anytime Repairing A*}
\newacronym{pca}{PCA}{Principal Component Analysis}
\newacronym{cpg}{CPG}{Central Pattern Generator}
\newacronym{wbc}{WBC}{Whole-Body Control}
\newacronym{pd}{PD}{Proportional-Derivative}
\newacronym{nmpc}{NMPC}{Nonlinear Model Predictive Control}
\newacronym{wbopt}{WBOpt}{Whole Body Optimization}

\newacronym{hc}{HC}{Hunt and Crossley's}
\newacronym{kv}{KV}{Kelvin-Voigt's}

\newacronym{wllsr}{WLLSR}{Weighted Linear Least Squared Regression}

\newacronym{mae}{MAE}{Mean Absolute Tracking Error}

\newacronym{lip}{LIP}{Linear Inverted Pendulum}

\newacronym{nn}{NN}{Neural Network}
\newacronym{drl}{Deep-RL}{Deep Reinforcement Learning}
\newacronym{td3}{TD3}{Twin Delayed Deep Deterministic Policy Gradient}
\newacronym{ddpg}{DDPG}{Deep Deterministic Policy Gradient}
\newacronym{rpe}{RPE}{Relative Percentual Error}
\newacronym{grl}{GRL}{Guided Reinforcement Learning}
\newacronym{e2e}{E2E}{End-to-end Reinforcement Learning}

\newcommand{\Rnum}{\mathbb{R}} 

\newcommand{\vect}[1]{\mathbf{#1}} 

\newcommand{\mat}[1]{\ensuremath{\begin{bmatrix}#1\end{bmatrix}}}	

\newtheorem{prob}{Problem}

\usepackage{resizegather}

\newcommand\BibTeX{{\rmfamily B\kern-.05em \textsc{i\kern-.025em b}\kern-.08em
T\kern-.1667em\lower.7ex\hbox{E}\kern-.125emX}}

\makeatletter
\newcounter{definition*}
\newenvironment{definition*}[1][htb]
{\renewcommand{\ALG@name}{Definition}
	\let\c@algocf\c@megaalgorithm
	\begin{algorithm*}[#1]%
	}{\end{algorithm*}}
\makeatother

\makeatletter
\newcounter{definition}

\makeatother

\definecolor{sfahmi_blue}{RGB}{0.19,0.51,0.74}
\definecolor{LightBlue}{RGB}{0.4,0.4,1}

\title{\LARGE \bf Efficient Reinforcement Learning for Jumping Monopods}
\author{Riccardo Bussola$^{1}$, Michele Focchi$^{1}$,  Andrea Del Prete$^{2}$, Daniele Fontanelli$^{2}$, Luigi Palopoli$^{1}$
	\thanks{$^1$ The authors are with the Dipartimento di Ingegneria and Scienza dell'Informazione (DISI), University of Trento, {\tt\small \href{mailto:name.surname@unitn.it}{name.surname@unitn.it}}}%
	\thanks{$^2$ The authors are with the Dipartimento di Ingegneria Industriale (DII), University of Trento,  {\tt\small \href{mailto:name.surname@unitn.it}{name.surname@unitn.it} }. We acknowledge the support of the MUR PNRR project FAIR - Future AI Research (PE00000013) and of the FSE-REACT-EU, PON Research and Innovation 2014-2020 DM1062 / 2021.} }
\begin{document}
	
	\thispagestyle{empty}
	\pagestyle{empty}
	\maketitle
	\begin{abstract}
		In this work, we consider the complex control problem of
		making a monopod reach a target with a jump. The monopod can jump in
		any direction and the terrain underneath its foot can
		be uneven. This is a template of a much larger class of problems,
		which are extremely challenging and computationally expensive to solve
		using standard optimisation-based techniques.
		Reinforcement Learning (RL) could be an interesting
		alternative, but the application of an end-to-end approach in which
		the controller must learn everything from scratch, is impractical. The solution
		advocated in this paper is to guide the learning process within an RL framework
		by injecting physical knowledge.
		This expedient brings to widespread benefits, such as a drastic reduction of the learning
		time, and the ability to learn and compensate for possible errors in the low-level controller executing the motion. 
		We demonstrate the advantage of our approach with respect to both optimization-based and end-to-end RL approaches. 
	\end{abstract}
	%
	\vspace{-0.1cm}
	\section{Introduction}
	\label{sec:introduction}
Legged robots have become a popular technology to navigate unstructured terrains,
but are complex devices for which control design is not trivial. Remarkable results
have been reached for locomotion tasks like walking and trotting~\cite{tamols22}.
Other tasks, like performing jumps, are more challenging because even small
deviations from the desired trajectory can have a large impact on the landing location and orientation \cite{roscia2023reactive}.
%
This problem has received some attention in the last few years.
A line of research has produced heuristic approaches
relying on physical intuitions and/or on simplified models to be used
in the design of controllers or planners \cite{park17, roy20}.
However, the hand-crafted motion actions produced by these approaches
are not guaranteed to be  physically implementable.
Another common approach is to use full-body numerical optimisation~\cite{nguyen2021contact}.
Very remarkable is the rich set of aerial motions produced by 
MIT Mini Cheetah in~\cite{chignoli2021online, garcia21, chignoli22}
using a centroidal momentum-based nonlinear optimisation.
A problem with optimisation-based approaches on high dimensional nonlinear problems is the 
high computational cost, which makes them unsuitable for a real-time implementation, 
especially to replan trajectories over a receding horizon.
Recent advances  \cite{mastalli22,nguyen2021contact} have brought to significant improvements in the efficiency of \gls{mpc} for 
jumping tasks. However, the price to pay is to 
introduce some artificial constraints such as fixing the contact sequence, 
the time-of-flight, or optimising the contact timings offline.


A third set of approaches is based on \gls{rl}.
The seminal work of Lillicrap~\cite{Lillicrap2015ContinuousCW} showed  
that a \gls{ddpg} algorithm combined with a Deep Q network could be successfully applied to learn end-to-end policies for a continuous 
action domain, using an actor-critic setting. 
In view of these results, several groups have then applied \gls{rl} to quadrupeds for locomotion tasks~\cite{Gehring2016, hwangbo2019, peng2020,Ji2022, Rudin2021}, and to \textit{in-place} hopping
legs~\cite{Fankhauser2013}.
As with most of model-free reinforcement
approaches, \gls{ddpg} requires a large number of training 
steps (on the order of millions) to find good solutions.
%
%
Other approaches~\cite{khadiv22, bellegarda2020robust,Grandesso2023} seek to improve the efficiency and the robustness of
the learning process by 
combining \gls{to} with \gls{rl}: they use the former to generate initial trajectories to bootstrap the exploration of alternatives made by the latter.

As a final remark, the efficiency and the robustness of the RL learning process is heavily affected by a correct choice of the action space~\cite{vanderpanne2017, bellegarda19}.
Some approaches require that the controller directly generates the torques \cite{Chen2022}, while others suggest that the controller should operate in Cartesian or joint space~ \cite{hwangbo2019, RL_solo12}.
%
%


\noindent \textbf{Paper Contribution}.
This work proposes an \gls{rl} framework capable of producing an omni-directional jump trajectory (from standstill) on \textit{uneven} terrain, computed within a few milliseconds, unlocking real-time usage with the
current controller rates (e.g. in the order of $kHz$).  
The main objective of this paper is to reduce the duration of the learning phase without sacrificing the system's performance.  
A reduced length of the learning phase has at least two indisputable advantages:
lowering the barriers for the access to this technology for
professionals and companies with a limited availability of computing power, 
and addressing the environmental concerns connected with the carbon footprint of learning technologies~\cite{henderson2020towards}.

Our strategy is based on the following ideas: first, learning is
performed in Cartesian space rather than in joint space so that the agent can more directly verify the effect of its
actions. 
Second, we notice that while the system is airborne, its final landing
point are dictated by simple mechanical laws (ballistic). Therefore, the learning process can focus solely on the \textit{thrusting} phase. 
Third, we know from biology~\cite{zador2019critique} that mammals are extremely effective in learning how to walk because of "prior" knowledge in their genetic background. 
This means that the learning process can be \textit{guided}
by an approximate knowledge of what the resulting motion should "look like". 
Specifically, we parametrise the thrusting trajectory
(i.e. from standstill to lift-off) for the \gls{com} with a ($3^{rd}$
order) Bezier curve. 
This choice of a Bezier curve to parametrize the action space is not uncommon in the literature \cite{petar2012}, \cite{Kim2021QuadrupedLO}, \cite{ji2022Bezier}, and in our case 
is motivated by the physical intuition that a jump motion, to exploit the full joint range, needs a "charging" phase to compress the legs
followed by an extension phase where the \gls{com} is accelerated both upwards and in the jump direction.  
Clearly, by making this restriction, we prevent the learning phase from exploring alternative options. 
However, as we will see in the paper, the final result is
very close to optimal and the system retains good generalisation abilities.

Our approach is based on TD3, a state-of-the-art \gls{drl} algorithm \cite{fujimoto2018td3}
trained to minimise a cost very similar to the ones typically used in optimal control.
The Cartesian trajectory generated by our \gls{rl} agent is translated into joint space via inverse kinematics,
and tracked by a low-level joint-space \gls{pd} controller with gravity compensation. 
Our results reveal that possible inaccuracies in the
controller can be learned and compensated by the \gls{rl} agent.
In this paper we do not focus on the landing phase, which we assume to be managed by a different controller, such as~\cite{roscia2023reactive}.
We compare our approach (that we will call  \gls{grl})  
with both  a baseline \gls{to} controller with a \textit{fixed} duration
of the thrusting phase, and a "standard" \gls{e2e} (which considers joint references as action space). 
In the first case, we achieved better or equal performance, 
and a dramatic reduction in the online computation time.
With respect to \gls{e2e} \textit{locomotion} approaches \cite{Chen2022, hwangbo2019},
we observed a substantial reduction in the 
number of episodes (without considering parallelization) needed to achieve good learning performances. Instead, an \gls{e2e} implementation specific for a  \textit{jump} motion (Section \ref{sec:e2e}), 
 did not provide satisfactory learning results.



 

\label{sec:outline}
The paper is organized as follows: Section \ref{sec:approach} 
presents the \gls{grl} approach, detailing the core components of the MDP (action, reward functions). 
Section \ref{sec:implementation} provides implementation details.  
In Section \ref{sec:results} we showcase our simulation results comparing 
with state-of-the-art approaches.
Finally, we draw the conclusions in Section \ref{sec:conclusions}.

	\vspace{-0.1cm}
	\section{Problem description and solution overview}
	\label{sec:approach}
 %

%
Simple notions of bio-mechanics suggest that legged animals execute
their jumps in three phases: 
1. \emph{thrust:} an initial compression is followed by an explosive extension of the limbs in order to gain sufficient momentum for the lift-off; the phase finishes when the foot leaves the ground; 
2. \emph{flight:} the body, subject uniquely to gravity, reaches an apex where the vertical \gls{com} velocity changes its sign and the robot adjusts its posture to prepare for landing; 
3. \emph{landing:} the body realizes a touch-down, which means that the foot establishes again contact with the ground. 
For the sake of simplicity, we consider a simplified and yet realistic setting:
a monopod robot, whose base link is sustained by passive
\textit{prismatic} joints preventing any change in its orientation (see Section \ref{sec:implemDetails}). 
The extension to a full quadruped (which requires considering also angular motions) is left for future work.
%
%
In this paper, \emph{we focus on the thrust
phase}. 
The flight phase is governed by the ballistic law. 
Let $\vect{c}_{tg}$ be the target location and let the \gls{com}
state at lift-off be $(\vect{c}_{lo}, \vect{\dot{c}}_{lo})$. 
After lift-off, the trajectory lies on the vertical plane containing $\vect{c}_{lo}$ and $\vect{c}_{tg}$. 
%
%
%
The set of possible landing \gls{com} positions are function of $\vect{c}_{lo}$,  $\vect{\dot{c}}_{lo}$ and are given by the following equation:
\begin{equation}
\label{eq:ballistic_eq}
    \begin{cases}
    \vect{c}_{tg,xy} = \vect{c}_{lo,xy}+\vect{\dot{c}}_{lo,xy}T_{fl}\\
    \vect{c}_{tg,z} =\vect{c}_{lo,z}+\vect{\dot{c}}_{lo,z}T_{fl}+  -\frac{1}{2}gT_{fl}^{2}
    \end{cases}    
\end{equation}

where $T_{fl} = (\vect{c}_{tg,xy}-\vect{c}_{lo,xy})/\vect{\dot{c}}_{lo,xy}$ is the flight time.
%
%
%
In this setting, our problem can be stated as follows:
\begin{prob}
  \label{prob:form}
    Synthesise a thrust phase that produces a lift-off configuration (i.e. \gls{com} position and velocity) that: 1. satisfies~\eqref{eq:ballistic_eq},
    2. copes with the potentially adverse conditions posed by the environment (i.e. contact stability, friction constraints), 
    3. satisfies the physical and control constraints.
\end{prob}

Nonlinear optimisation is frequently used for 
similar problems. However, it has two important limitations that discourage its application in
our specific case: 1. the computation requirements are very high, complicating both the real--time 
execution and the use of low--cost
embedded hardware, 2. the problem is strongly non-convex, which can lead the solvers to local minima.
%
\vspace{-0.1cm}
\subsection{Overview of the approach}
In this work, we use \gls{rl} to learn optimal joint  trajectories to realise a jump motion, 
that is then tracked by a lower-level controller.
We adopted the state-of-the-art \gls{td3}, a
\gls{drl} technique.  
This algorithm is based on the well-known Actor-Critic architecture. 
The main idea behind \gls{td3} is to adopt two Deep \gls{nn} to approximate the policy,
represented by the Actor, and two Deep \gls{nn} to approximate the action-value function, represented by the Critic.
\begin{figure}[tbp]
\centering
\includegraphics[width=0.85\columnwidth]{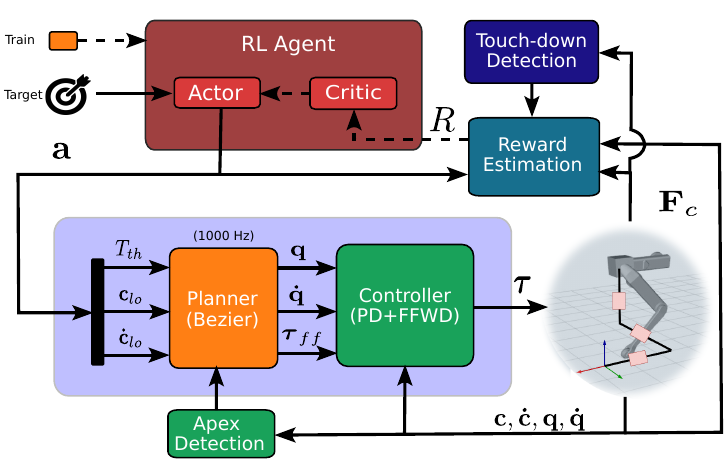}
\caption{\small  Diagram of the \gls{rl} Framework. The framework is split into two levels: 
the RL agents  and the planner.  
The RL agent  produces an action for the planner, 
based on a desired target. This computes a Bezier reference curve that is 
mapped into joint motion via inverse kinematics and tracked by 
the \gls{pd} controller that provides the joint torques to feed the  robot.
During the training, at the end of each  episode a 
reward is computed and fed back to the \gls{rl} agent.} 
\vspace{-0.5cm}
\label{fig:framework}
\end{figure}
%
%
%
%
Our \gls{rl} pipeline is  depicted in Fig. \ref{fig:framework}.
%
%
%
%
	%
	\vspace{-0.1cm}
	\section{Design and implementation of the RL agent}
	\label{sec:implementation}
	Based on the discussion in the previous section, the thrust phase is
characterised by the lift-off position $\vect{c}_{lo}$ and velocity
$\vect{\dot{c}}_{lo}$ and by the thrust time $T_{th}$ which is the time spent to reach the lift-off
configuration from the initial state.  
The state of the environment is defined as ($\vect{c}$, $\vect{c}_{tg}$) where $\vect{c} \in \Rnum^3$ is the \gls{com}
position and $\vect{c}_{tg} \in \Rnum^3$ the \gls{com} at the landing location (\textit{target}). 
The objective of the \gls{rl} agent is to find the jump parameters ($\vect{c}_{lo}$, $\vect{\dot{c}}_{lo}$, $T_{th} \in \Rnum$), that minimise the landing error at touch-down $\Vert \vect{c} - \vect{c}_{tg}\Vert$ while satisfying the physical constraints.
Our jumping scenario thought can be seen as a \textit{single-step} trajectory where the only action performed leads always to the end state.

\subsection{The Action Space}

%
The dimension of the action space has a strong impact on the performance of the \gls{rl} algorithm.
Indeed, a \gls{nn} with a smaller number of outputs is usually faster to train. 
What is more, a smaller action space reduces the complexity of the mapping, speeding up the learning process.
%

A first way to reduce the complexity of the action space is 
by expressing $\vect{c}_{lo}$ and $\vect{\dot{c}}_{lo}$ in spherical coordinates. 
Because of the peculiar nature of a jump task, the trajectory 
lies in the plane containing  the \gls{com} $\vect{c}$ 
and its desired target location  $\vect{c}_{tg}$. Hence, 
the yaw angle $\varphi$ remains constant ($\bar{\varphi}$) throughout the flight and
we can further restrict the coordinates to a convex bi-dimensional space:
\begin{align}
\small
\begin{cases}
   \vect{c}_{lo,x} = r~\cos(\theta)\cos(\bar{\varphi})  \\
   \vect{c}_{lo,y} = r~\cos(\theta) \sin(\bar{\varphi})  \\
   \vect{c}_{lo,z} = r~\sin(\theta) 
\end{cases} 
\begin{cases}
    \vect{\dot{c}}_{lo,x} = r_v~\cos(\theta_v)\cos(\bar{\varphi}) \\
    \vect{\dot{c}}_{lo,y} = r_v~\cos(\theta_v)\sin(\bar{\varphi})  \\
    \vect{\dot{c}}_{lo,z} = r~\sin(\theta_v) 
\end{cases}
\label{eq:possible_lo_pos}\raisetag{2\normalbaselineskip}
\end{align}

\begin{figure}[tbp]
\centering
\includegraphics[width=0.9 \columnwidth]{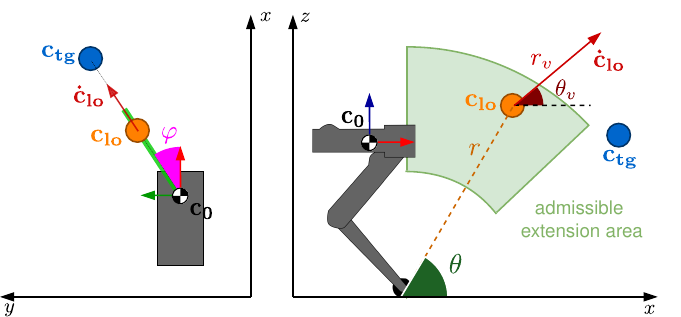}
\caption{\small Action parametrization and its bounds. On the left the top view, on the right the side view of the jumping plane.}
\label{fig:simpification_draw}
\vspace{-0.5cm}
\end{figure}

As shown in Fig.~\ref{fig:simpification_draw}, the lift-off position $\vect{c}_{lo}$ is identified by: the radius $r$ (i.e., the maximum leg extension), the yaw angle $\varphi$  and the pitch angle $\theta$. 
Likewise, the lift-off velocity $\dot{c}_{lo}$, is described by  its  magnitude $r_{v}$, and the pitch angle $\theta_{v}$ with respect to the ground.
%
%
Therefore, by using this assumption, we have reduced the dimension of the action space from $7$ to $5$: 
$\vect{a} = (T_{th}, r, \theta, r_v, \theta_v) \in \Rnum^5$.

The action space can be further restricted by applying some domain knowledge. 
The radius $r$ has to be smaller than a value $r_{max}$ ($0.32$ $m$) to prevent boundary singularity due to 
over-extension, and greater than a value $r_{min}$($0.25$ $m$) to avoid complete leg retraction. 
The bounds on the velocity $\dot{c}_{lo}$,  represented by $r_{v} \in \left[0.1,4\right]$ $m/s$, 
and $\theta_{v} \in \left[\frac{\pi}{6},\frac{\pi}{2}\right]$ $rad$, and the bounds on pitch angle 
$\theta \in \left[\frac{\pi}{4},\frac{\pi}{2}\right]$  $rad$ are set to rule out jumps that involve excessive 
foot slippage and useless force effort. 
Specifically, restricting to a positive  $\theta_{v,min}$ 
ensures  a non-negligible vertical component for the velocity, while
bounding  $\theta_v$ to the positive quadrant  secures that the 
lift-off velocity be oriented  "toward" the target.
%
%
\subsubsection{Trajectory Parametrisation in Cartesian Space}


Our strategy to tackle the problem of  generating a compression-extension trajectory for the leg,
to achieve a given lift-off configuration $\vect{c}_{lo}$, is based on two important choices:
1. making the RL agent learn the trajectory in Cartesian Space and then finding the joint trajectories through inverse kinematics, 2. restricting the search of the Cartesian space evolution to  curves generated by known parametric functions. This has the aim to reduce the search space and simplifying convergence.


The analytical and geometric properties of $3^{rd}$ order  
Bezi\'er curves make them a perfect fit for our problem.
A $3^{rd}$  order Bèzier curve is defined by \textit{four} control points. In our case, the first and the final points
are constrained to be the initial and final \gls{com} positions, respectively.
The derivative of a  $3^{rd}$ degree Bèzier curve is itself a Bèzier curve of $2^{nd}$ degree with 3 control points defined as $3(\vect{P}_{i+1}-\vect{P}_{i})$.
The curve domain is defined only in the normalised time interval: $t \in  \mat{0, 1}$. 
%
Defining the following Bernstein polynomials: 
\begin{align}
    &\boldsymbol{\eta}(t) = \mat{(1-t)^3 & 3(1-t)^2t & 3(1-t)t^2 & t^3} ^T  \\
    & \boldsymbol{\dot{\eta}}(t) = \mat{(1-t)^2 & 2(1-t)t & t^2} ^T
\end{align}
we can compactly write the Bezi\'er curve as function of its $\vect{P}_i \in \Rnum^3$ control points:
%
\begin{align}
\label{eq:bezier}
\vect{c} =  
\mat{\vect{P}_{0} & \vect{P}_{1} & \vect{P}_{2} & \vect{P}_{3}}   \boldsymbol{\eta}(t) 
\end{align}
Since we are considering an execution time $T_{exe} 
\in \mat{ 0 , T_{th}}$ and  $t = \frac{T_{exe}}{T_{th}}$, then  the derivative writes:
%
\begin{align}
\label{eq:bezier_dot}
\vect{\dot{c}} = \frac{1}{T_{th}}       
\mat{\vect{P}'_{0} & \vect{P}'_{1}     &  \vect{P}'_{2}} \boldsymbol{\dot{\eta}}(t)
\end{align}
From the definition of the curve \eqref{eq:bezier} and its derivative 
\eqref{eq:bezier_dot}, we can compute the control points $\vect{P}_i$ 
by setting the boundary conditions of  the initial/lift-off  \gls{com} position 
$\vect{c}_{0}$, $\vect{c}_{lo}$  and initial/lift-off 
\gls{com} velocity  $\vect{\dot{c}}_{0}$,$\vect{\dot{c}}_{lo}$ in \eqref{eq:bezier_cp}. 
%
%
%
%
\begin{equation}
\small
\begin{cases}
    \begin{aligned}
        {\vect{P}}'_{0} &= \frac{3}{T_{th}}(\vect{P}_{1}-\vect{P}_{0})\\
                        &= \vect{\dot{c}}_{0}= \vect{0}\\
        {\vect{P}}'_{1} &= \frac{3}{T_{th}}(\vect{P}_{2}-\vect{P}_{1})\\
        {\vect{P}}'_{2} &= \frac{3}{T_{th}}(\vect{P}_{3}-\vect{P}_{2})\\
                        &=\vect{\dot{c}}_{lo}
   \end{aligned}
\end{cases}
\begin{cases}
    \begin{aligned}
        \vect{P}_{0} &= \vect{c}_{0}\\
        \vect{P}_{1} &= \frac{T_{th}}{3}{\vect{P}}'_{0}+\vect{P}_{0}\\
                     &=\frac{T_{th}}{3}\vect{\dot{c}}_{0}+\vect{c}_{0}\\
        \vect{P}_{2} &= -\frac{T_{th}}{3}{\vect{P}}'_{2}+\vect{P}_{3}\\
                     &=-\frac{T_{th}}{3}\vect{\dot{c}}_{lo}+\vect{c}_{lo}\\
        \vect{P}_{3} &= \vect{c}_{lo}\\
     \end{aligned}
\end{cases}
\label{eq:bezier_cp}
\end{equation}
\vspace{-0.3cm}
%
\subsection{A physically informative reward function}
In RL, an appropriate choice of the reward function is key to the final outcome.
Furthermore, we can use the reward function as a means to inject
prior knowledge into the learning process. 
In our case, the reward function was designed to penalise the violations of the physical constraints while giving a positive reward to the executions that make the robot land in proximity of the target point.
%
%
The constraints that must be enforced throughout the whole thrust phase 
are called  \textit{path} constraints. To transform the violations into costs, we introduce 
a linear activation function $A(x, \underline{x}, \bar{x})$ of the evaluated constraint, as function of the value $x$ and its lower and upper limits $\underline{x}$,~$\bar{x}$:
\begin{equation*}
    A(x,\underline{x}, \bar{x}) = \left | \min(x-\underline{x},0)+\max(x-\bar{x},0)\right |
  \end{equation*}
The  output of the activation function is zero if the value is 
in the allowed range, the violation otherwise.


\noindent \textbf{Physical feasibility check}:
Before starting each episode, we perform a sanity check on the action $\vect{a}$: if the vertical velocity is not sufficient to reach the target height, we abort the simulation returning
a high penalty cost $C_{ph}$.
This can be computed by obtaining the time to reach the apex $T_{fup} = \vect{\dot{c}}_{{lo},z}/g$ 
%
%
and substituting it in the ballistic equation:
\begin{align}
    \bar{\vect{c}}_z(T_{fup}) &= \vect{c}_{{lo},z} + \vect{\dot{c}}_{{lo},z}T_{fup} + \frac{1}{2}(-g)T_{fup}^2  
\label{eq:apex}
\end{align}
this results in $\bar{\vect{c}}_z(T_{fup}) = \vect{c}_{{lo},z} +  \frac{1}{2}\frac{\vect{\dot{c}}_{{lo},z}^2}{g}$ which is the apex elevation. If $\vect{c}_{tg,z}>\bar{\vect{c}}_z(T_{fup})$, the episode is aborted.

\noindent \textbf{Unilaterality constraint}: 
in a legged robot, a leg can only push on the ground and not pull. 
This is because the component of the force $\vect{F}$ along the contact normal ($Z$ for flat terrains), must be positive.\\ 
\textbf{Friction constraint}: To avoid slippage, the tangential component  of the contact 
force $\left \|\vect{F}_{x,y} \right \|$ is bounded by the foot-terrain friction coefficient $\mu$: $\left \|\vect{F}_{x,y} \right \| \leq \mu \vect{F}_{z}$.\\
\textbf{Joint range and torque constraints}: the three joint $\bar{q}_{i}$  kinematic limits must not be exceeded. Similarly, each of the joint actuator torque $\bar{\tau}_{i}$ limits must be respected.\\
\textbf{Singularity constraint}: 
the singularity constraint 
avoids the leg being completely stretched.
During the thrusting phase 
the \gls{com} $\vect{c}$ must stay in the hemisphere of radius equal to the maximum leg extension.  
This condition prevents the robot from getting close to 
points with reduced mobility that produce high joint velocities in the inverse kinematic computation. 
Even though this constraint is enforced by construction in the action generation,
the actual trajectory might still violate it due to tracking inaccuracies.  
If a  singular configuration is reached, the episode is interrupted and a high cost is returned.
%
The costs caused by the violation of path constraints are evaluated 
for each time step of the thrust phase and accumulated into the feasibility cost $C_f$.
%
In addition to these path constraints, we also want to 
take into account the error between actual and desired 
lift-off state. 
This penalty $C_{lo}$ encourages lift-off configurations that are better tracked by the controller. 
Another penalty $C_{td}$ is introduced when an episode does not result in a touchdown. 
This is done to force in-place jumps, and avoid the robot to stay stationary.
%
The positive component of the reward function is the output 
of a nonlinear \textit{landing target} reward function, 
which evaluates how close the \gls{com} arrived to the desired target. 
This reward grows exponentially when this distance approaches zero:
\begin{equation}
    R_{lt}(\vect{c},\vect{c}_{tg})= \frac{\beta}{k \left \|\vect{c}-\vect{c}_{tg}\right \| + \epsilon},
\label{eq:target_function}
\end{equation}
where $k$ is a gain to encourage jumps closer to the target position, and $\beta$ is an 
adjustable parameter to bound the max value of $R_{lt}$ and scale it. 
An infinitesimal value $\epsilon$ is added at the denominator to avoid division by zero.
%
Hence, the total reward function is:
\begin{equation}
R = 1_{\Rnum^+} \left[ R_{lt}(\vect{c},\vect{c}_{tg})- \sum_{i=0}^{n_c}C_{i} \right]
\end{equation}
with $n_c = 8$, and where $C_i$ are the previously introduced feasibility costs. 
We  decided to perform \textit{reward shaping}\cite{grzes2017reward}, by clamping the total reward to $\Rnum^{+}$ by mean of an indicator function $1_{\Rnum^+}$. This aims to promote the actions that induce constraint satisfaction.
%

\subsection{Implementation details}
\label{sec:implemDetails}
The training of the \gls{rl} agent and the sim-to-sim validation of the 
learned policy was performed on top of a Gazebo simulator. 
Because we are considering  only translational motions, we modelled a 3 \gls{dofs} monopod 
with three passive prismatic joints attached at the base.
These prismatic joints constrain the robot base's movements to  planes parallel to the ground.
For the sake of simplicity, we also considered 
the landing phase under the responsibility
of a different controller (e.g., see~\cite{roscia2023reactive}). 
Our interest was simply on the touch-down event, which is checked
by verifying that the contact force  exceeds a positive threshold $f_{th}$. 
%
%
Therefore, the termination of the episode is determined by the 
occurrence of three possible conditions: execution timeout, singularity, or 
touch-down event.


The control policy (\textit{default \gls{nn}}) is implemented as a neural
network that takes the \textit{state} as an input, and outputs the \textit{actions}.
The \gls{nn} is  a multi-layer perceptron
with three hidden layers of sizes 256, 512 and 256 with
ReLU activations between each layer, and with \textit{tanh} 
output activation to map the output between -1 and 1.
A low-level \gls{pd} plus gravity compensation controller 
generates the torques that are sent to the Gazebo simulator at 1 kHz.
The joint reference positions at  the lift-off are reset to the initial 
configuration $\vect{q}_0$ to enable the natural retraction of the leg and avoid stumbling.
Landing locations at different heights are achieved by making a 5x5 cm platform appear
at the desired landing location only at the apex moment. \footnote{Making the platform appear only at apex is needed for purely vertical jumps, because it avoids impacts with the platform during the trusting phase.}
The impact of the foot with the platform determines the touch-down moment and the consequent termination of the \textit{episode}.  
%
%
To train the \gls{rl} agent, the interaction with the simulation environment is needed. 
The communication between the  planner component  and the  Gazebo simulator is 
managed by the Locosim framework \cite{focchi2023locosim}. To interact with the planner, and consequentially, with the environment, we developed a ROS 
node called \textit{JumplegAgent}, where we implemented the \gls{rl} agent. The code is available at \footnote{Source code available at \href{https://github.com/mfocchi/jump_rl}{https://github.com/mfocchi/jump\_rl}}.
%
%
During the initial stage of the training process, the action is randomly generated to 
allow for an initial broad exploration of the action space for $N_{exp}$ episodes. 
%
%
%
	%
	\vspace{-0.cm}
	\section{Simulation Results}
	\label{sec:results}
	In this section, we discuss some simulation results that
show the validity of the proposed approach and compare it with
state-of-the-art approaches.
\begin{figure*}[tp]
    \includegraphics[width=\textwidth]{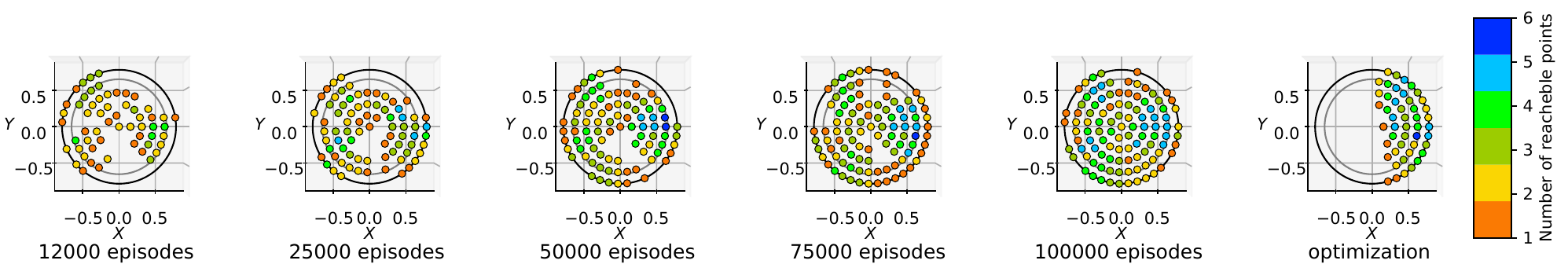}
    \caption{\small  Top-view of the feasibility region: (1-5) for different number of episodes of the training phase  (the number of reachable points is computed for each $X$,$Y$ pair) and (right) in the case of the baseline \gls{fddp}.}
    \label{fig:top_view_feasible_filtered}
\end{figure*}
%
%
%
%
We used a computer with the following hardware specifications:
CPU AMD Ryzen 5 3600, GPU Nvidia GTX1650 4GB, RAM 16 GB DDR4 3200 MHz.
During training we generated  targets inside a 
predefined \textit{training} region. This samples are generated \textit{randomly} 
inside a cylinder centered on the robot's initial \gls{com}, with a radius 
from 0 to $0.65$~m and a height from $0.25$~m to $0.5$~m. 
The size was selected to push the system to its performance limits. 
%
%
The parameters of the robot, controller and simulation are presented in Table \ref{tab:robot_params}.
\begin{table}[tbp]
\begin{center}
    \vspace{-0.5cm}
\caption{Controller, Planner, Simulator parameters}
\begin{tabular}{c| p{3.5cm} | c } 
    \hline
    \textbf{Variable} & \textbf{Name} & \textbf{Range}\\
    \hline    \hline
     $m$ &  Robot Mass [$kg$] & 1.5\\
    \hline
     P & Proportional gain & 10 \\
    \hline
     D & Derivative gain & 0.2 \\
    \hline
     $\vect{q}_0$ & Nominal configuration & \mat{0&-0.75&1.5} [$rad$] \\
    \hline
     $dT$ & Simulator time step [$s$] & 0.001\\
    \hline
     $\tau_{max}$ & Max torque [$Nm$] & 8 \\
    \hline
     $f_{th}$ & Touch-down force th. [$N$] & 1 \\
    \hline
    $N_{exp}$     &  Num. of  expl. steps    & 1280 (\gls{grl}), 10e4 (\gls{e2e})\\
    \hline
    $b_{s}$     &  Batch size    & 256 (\gls{grl}), 512 (\gls{e2e})\\
    \hline 
    $n_{exp}$     &  Expl. noise    & 0.4 (\gls{grl}), 0.3 (\gls{e2e})\\
    \hline 
    $tg_{rep}$     &  Landing target repetition    & 5 (\gls{grl}), 20 (\gls{e2e})\\
    \hline 
    $N_{train}$     &  Training step interval  & 1 (\gls{grl}), 100 (\gls{e2e})\\
    \hline 
    \hline
\end{tabular}
    \vspace{-0.5cm}
\label{tab:robot_params}
\end{center}
\end{table}


\subsubsection{Nonlinear Trajectory Optimisation}
The first approach to compare with is a standard optimal control strategy
based on \gls{fddp}. 
\gls{fddp} is one of the most efficient optimisation algorithms for whole-body control \cite{Budhiraja2018}
because it takes full advantage of the intrinsic sparse structure of the optimal control problem.
The \gls{fddp} solver is implemented with the optimal control library Crocoddyl \cite{crocoddyl20}
and uses the library Pinocchio \cite{carpentier2019pinocchio} to enable fast computations of costs, dynamics, and their derivatives.

For the problem at hand, we discretised the trajectory into $N$ successive knots with a timestep $dT = 0.001$ $s$ to ensure high precision. 
As decision variables we chose the joint torques.
We split the jump in three phases: thrusting, flying, and landed.
The constraints in \gls{fddp} are encoded as soft penalties in the cost. We encoded friction cones, and tracking of foot locations and velocities at the thrusting/landed stages, respectively. We added a tracking cost for the \gls{com} reference during the flying phase to encourage the robot to lift-off. We regularized control inputs and states throughout the horizon.

\subsubsection{End-to-end RL}
\label{sec:e2e}
At the opposite side of the spectrum of optimal control 
is the application of approaches entirely based on Deep Learning, 
i.e., using \gls{rl} end-to-end \textit{without} injecting any prior domain knowledge.
%
%
The \gls{rl} agent sets joint position references to a low-level (PD)+gravity compensation controller.
The use of a PD controller allows the system to inherit
the stabilisation properties of the  feedback controller, but at the same time, it allows for 
explosive torques (by regulating the references to have a bigger error w.r.t. to the actual positions).
We query the action 
until the apex moment because we need to have set-points for the joints also when the leg is air-borne. After apex, we set the default configuration $\vect{q}_0$ for the landing.
%
%
%
%
To be more specific, instead of directly setting the joint references $\vect{q}^d$, the control policy produces as action $\vect{a}$ joint angle deviations $\vect{\tilde{q}} \in \Rnum^3$  w.r.t.  to the nominal joint angle configuration $\vect{q}_0$.

As suggested by  \cite{Gangapurwala2022, Zhao2023} to ease the learning, we run at a frequency that is 1/5th than the controller one (1 $kHz$).
We aggregate the reward for each control loop iteration 
and perform a training every $N_{train}=100$ queries. 
We terminate the episode if: 1) a touchdown is detected, 2) the robot has fallen (i.e. the base link close to the ground), 3) we reach singularity 4) a timeout of 2.5 $s$ is reached. 
We include in the state \gls{com} and joint position/velocity.
Since the domain is not changing (and  the state is Markovian), 
augmenting the state with the history of some  past samples \cite{Aractingi2022, hwangbo2019}
was not necessary, therefore we tried to keep the problem dimensionality as low as possible. 
%
%
Hence, the observation state becomes 
($\vect{c}$, $\vect{q}$, $\vect{\dot{c}}$, $\vect{\dot{q}}$,  $\vect{c}_{tg}$).  
As in the case of the \gls{grl} approach, the initial 
state at the start of each episode is set at the
nominal joint pose ($\vect{c}_0, \vect{q}_0$), with zero velocity.
We also encourage \textit{smoothness} by penalizing the quantity $\vect{\tilde{q}}^{j} - \vect{\tilde{q}}^{j-1}$. Because of the different units, to have better conditioning in the gradient of the \gls{nn}, we scale each state variable against its range. 
With respect to the \gls{grl} approach we increase the batch size to 512 to collect more observations for training. We provide the  feasibility rewards $C_i$ at each loop as \textit{differential} w.r.t. the previous loop. This approach is meant to ease learning while converging to the same solution \cite{Jeon2023}. 
%
%
At the end of each episode, we strongly penalize the lack of 
a touch-down event, and use the same target function \eqref{eq:target_function} to encourage landing close to the target. 
The fact that we provide this quantity at every step enables us to achieve an informative reward \cite{Jeon2023} also \textit{before} the touchdown.
Following the \textit{curriculum learning} idea \cite{Bengio2009}, we gradually increase the difficulty of the jump  enlarging the bounds of the training region (where the targets are sampled) in accordance with the number of episodes and the average reward.

\vspace{-0.1cm}
\subsection{Policy performance: the feasibility region}
We tested the agent in \textit{inference} mode, for omni-directional jumps at
different heights, for  726 target positions \textit{uniformly} spaced  on a 
grid  (\textit{test region}) of the same shape of the training region.  
The \textit{test region}  is 20$\%$ bigger than
the training region, in order to demonstrate the generalisation
abilities of the system. 

The policy was periodically evaluated on the \textit{test region} set
in order to assess the evolution of the models stored during the training phase. 
To measure the quality of a jump, we used the \gls{rpe}, 
which we define as the distance between the touch-down and the 
desired target point, divided by the jump length.
%
%
%
%
The feasibility region represents the area where the agent is capable of an accurate landing, i.e. \gls{rpe} $\le  10\%$. 

\subsubsection{Performance baseline: Trajectory Optimisation}
We compared the approach with the baseline \gls{fddp} approach 
repeating the same optimisation for all the points in the 
\textit{test region} without changing the costs'weights and limiting the number of iterations to $500$.  
For optimal control, the average computation time was $17$ $s$ 
for back jumps and $7.6$ s for front jumps while a single 
evaluation of the \gls{nn} requires only $0.7$ $ms$.
Fig. \ref{fig:top_view_feasible_filtered} (right) shows that a 
reasonable accuracy  is obtained for landing locations in the 
front of the robot, while \gls{fddp} behaves poorly   for locations  in the back of the robot.
%
Computing the mean \gls{rpe} separately for the back region 
and the front region, we obtained an accuracy of 52~$\%$ and  
16.5$\%$, respectively.
\subsubsection{Performance of the GRL approach}
\begin{figure}[tbp]
    \centering
    \includegraphics[width=0.8\linewidth]{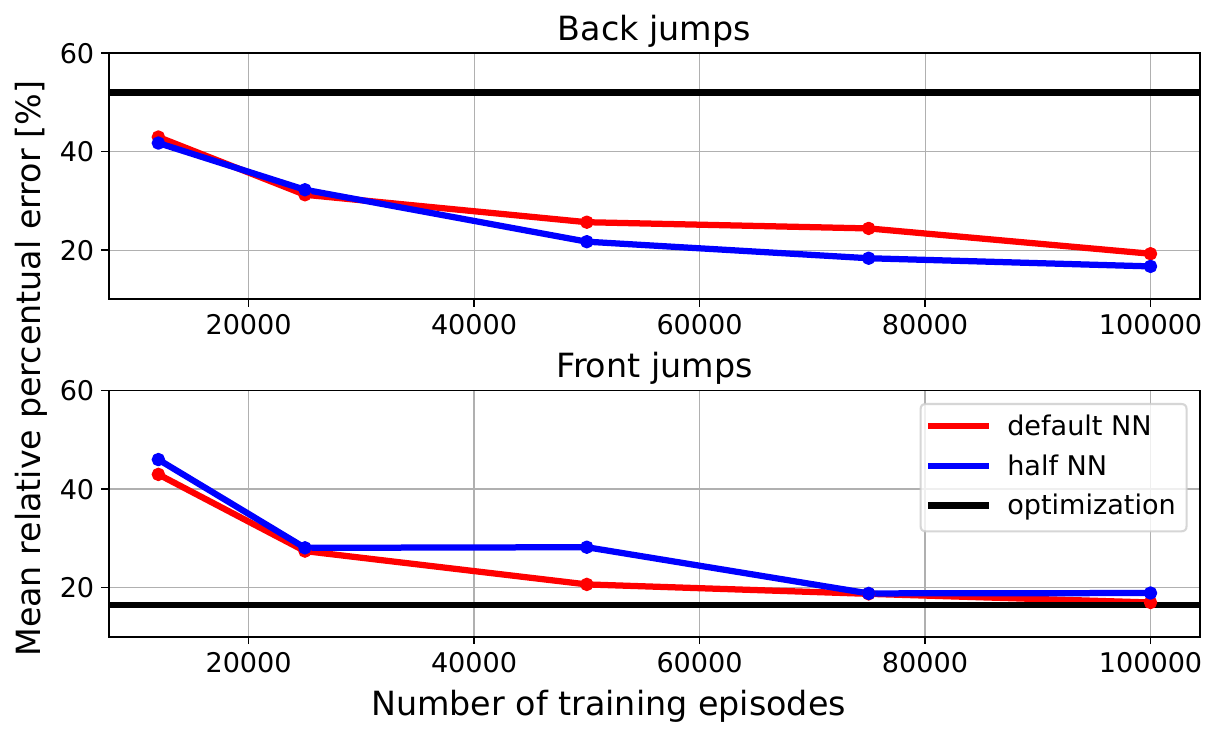}
    \caption{\small  Plot of the average \gls{rpe} as a function of the number of training episodes.}
    \label{fig:avg_error}
    \vspace{-0.5cm}
\end{figure}

We repeated the simulations on the test region using the \gls{grl} approach with the \textit{default} \gls{nn} and with a \gls{nn} where we halved the number of neurons in each hidden layer (\textit{half \gls{nn}}).
Fig. \ref{fig:avg_error} shows that, in both cases,  with our \gls{grl} approach 
the \gls{rpe} decreases (i.e.  accuracy increases)
monotonically  with the number of episodes. In the front case,  
going from approximately 40$\%$  to 16$\%$  for 100k episodes. A satisfactory 
level (i.e. \gls{rpe} 20$\%$ for front jumps) is already achieved  after 50k episodes. 
All the feasibility constraints turn out to be mostly  satisfied  after 10k episodes. 
The figure also shows that the \gls{grl} approach always outperforms the standard optimisation 
method in terms of jump accuracy in the case of back jumps,
achieving a comparable accuracy for front jumps. 
Halving the number of neurons, the two models behave with similar 
accuracy, showing that the \gls{rl} model could ideally be further simplified.
%
%
From Fig. \ref{fig:top_view_feasible_filtered} we can observe  that the feasibility 
region expands with the number of training epochs.
%
In the same figure we can see that the \textit{test region} (black cylinder) is bigger than the  region where we trained the \gls{nn} (gray cylinder), 
demonstrating its extrapolation capabilities.
It is important to remark that the \gls{grl} approach is also capable, to some extent, to learn the dynamics and compensate the tracking inaccuracies of the underlying low-level controller.  
This is shown in the accompanying video \footnote{\href{https://www.dropbox.com/scl/fi/89hv8cfsrqd3nyx34p9kd/bussola23icra.mp4?rlkey=qzl1asgna4aagohieviqxdwc4&dl=0}{Link to the accompanying video}} by the \textit{purple} ball that represents the \textit{ideal} landing location (i.e. if the \gls{com} lift-off velocity associated to each action  was perfectly tracked). 
 The \textit{purple} ball is different from the target location (\textit{blue} ball) because of tracking inaccuracies but the agent learned to provide a lift-off velocity that  compensates for these, managing to reach accurately the desired location (blue ball). 
In the same video we show how that the quality of the jumps  steadily  improves with the  number of training episodes.
%
\subsubsection{Performance baseline: end to end RL}
We did not  achieve satisfactory results with the \gls{e2e} approach.
The reward function had an erratic behaviour during the training and even after 1M of training steps there were only a few targets out of $726$ where the algorithm managed to attain an error below $10$ $\%$.  More details are reported in the accompanying video.
\vspace{-0.2cm}

	\vspace{-0.0cm}
	\section{Conclusions}
	\label{sec:conclusions}
	%
In this work we proposed a guided \gls{rl} based strategy to perform
\textit{omni-directional} jumps on uneven terrain with a legged robot.
Exploiting some domain knowledge and taking a few assumptions on the
shape of the jump trajectory, we have shown that, in a few
thousand episodes of training, the agent obtains the ability to jump
in a big area maintaining high accuracy while reaching the boundary of his
performances (deriving from the the physical limitations of the
machine).  The approach manages also to learn and proficiently
compensate from the tracking inaccuracies of the low-level controller.
The proposed approach is very efficient (it requires a small number of
training episodes to reach a good performance),  it achieves a good generalisation (e.g.,
by executing jumps in a region 20\% larger than the one used for
training), and it outperforms a standard end-to-end RL that resulted
not able to learn the jumping motion. 
Compared to optimal control, the \gls{grl} approach 1) achieves the 
same level of performance in  front jumps, but is also able 
to perform backward jumps (optimal control is not) 
2) requires several orders of magnitude lower computation time.
%
%
%
%
%
In the future, we plan to extend the approach to a full quadruped robot, considering not only linear 
but also angular motions. Leveraging the angular part we can build a framework that is able to  perform  
a variety of jumping motions (e.g. twist, somersault, barrel jumps) on inclined  surfaces.
We are also seeking ways to improve robustness by including robot 
non idealities in the learning phase and to speed up the training phase by leveraging parallel computation.

%
%
%
%
%

	%
	
	\bibliographystyle{style/IEEEtran}
	\bibliography{references/references}
\end{document}